\documentclass[11pt,a4paper]{article}
\usepackage[hyperref]{emnlp-ijcnlp-2019}

\usepackage{times}
\usepackage{latexsym}
\usepackage{amsmath} 
\usepackage{cleveref}
\usepackage{url}
\usepackage{booktabs}
\usepackage{float}
\usepackage{graphicx}
\usepackage{multirow}
\usepackage{arydshln}

\usepackage[warn]{textcomp}   

\aclfinalcopy 

\title{On the Importance of the Kullback-Leibler Divergence Term \\in Variational Autoencoders for Text Generation}

\author{Victor Prokhorov$^\clubsuit$, Ehsan Shareghi$^\clubsuit$, Yingzhen Li$^\spadesuit$ \\\bf{Mohammad Taher Pilehvar}$^{\clubsuit\diamondsuit}$, Nigel Collier$^\clubsuit$ \\
  $^\clubsuit$~Language Technology Lab, DTAL, University of Cambridge\\
  $^\spadesuit$Microsoft Research Cambridge ,$^\diamondsuit$Tehran Institute for Advanced Studies\\
  {\tt  $^\clubsuit$\{vp361, es776, mp792, nhc30\}@cam.ac.uk}\\
  {\tt $^\spadesuit$Yingzhen.Li@microsoft.com}}

\date{}

\begin{document}
\maketitle
\begin{abstract}
Variational Autoencoders (VAEs) are known to suffer from learning uninformative latent representation of the input due to issues such as approximated posterior collapse, or entanglement of the latent space. We
impose an explicit constraint on the Kullback-Leibler (KL) divergence term inside the VAE objective function. While the explicit constraint naturally avoids posterior collapse, we use it to further understand the significance of the KL term in controlling the information transmitted through the VAE channel. Within this framework, we explore different properties of the estimated posterior distribution, and highlight the trade-off between the amount of information encoded in a latent code during training, and the generative capacity of the model.~\footnote{The code is available on  \url{https://github.com/VictorProkhorov/KL_Text_VAE}}


\end{abstract}

\section{Introduction}
Despite the recent success of deep generative models such as Variational Autoencoders (VAEs)~\cite{Kingma2013AutoEncodingVB} and Generative Adversarial Networks (GANs)~\cite{DBLP:conf/nips/GoodfellowPMXWOCB14}  in different areas of Machine Learning, they have failed to produce similar generative quality in NLP. In this paper we focus on VAEs and their mathematical underpinning to explain their behaviors in the context of text generation.

The vanilla VAE applied to text \cite{DBLP:journals/corr/BowmanVVDJB15} consists of an encoder (inference) and decoder (generative) networks: Given an input $x$, the encoder network parameterizes $q_\phi(z|x)$ and infers about latent continuous representations of $x$, while the decoder network parameterizes $p_\theta(x|z)$ and generates $x$ from the continuous code $z$. The two models are jointly trained by maximizing the Evidence Lower Bound (ELBO), $\mathcal{L}(\theta, \phi; x,z)$:
\begin{equation}\label{elboobjective}
\setlength{\abovedisplayskip}{3pt}
\setlength{\belowdisplayskip}{3pt}
\big \langle \log p_\theta({x}|{z}) \big \rangle_{q_\phi({z}|{x})}  -  \vphantom{\big \langle \log p_\theta({x}|{z}) \big \rangle_{{z} \sim q_\phi({z}|{x})}} D_{KL}\big(q_\phi({z}|{x}) || p({z})\big)
\end{equation}
where the first term is the reconstruction term, and the second term is the Kullback-Leibler~(KL) divergence between the posterior distribution of latent variable $z$ and its prior $p({z})$ (i.e., $\mathcal{N}(0,I)$). The KL term can be interpreted as a regularizer which prevents the inference network from copying ${x}$ into ${z}$, and for the case of a Gaussian prior and posterior has a closed-form solution.


With powerful autoregressive decoders, such as LSTMs, the internal decoder's cells are likely to suffice for representing the sentence,  leading to a sub-optimal solution where the decoder ignores the inferred latent code ${z}$. This allows the encoder to become independent of $x$, an issue known as posterior collapse ($q_\phi({z}|{x})\approx p({z})$) where the inference network produces uninformative latent variables.
Several solutions have been proposed to address the posterior collapse issue: (i) Modifying the architecture of the model by weakening decoders \cite{DBLP:journals/corr/BowmanVVDJB15, DBLP:journals/corr/MiaoYB15, DBLP:journals/corr/YangHSB17, semeniuta-etal-2017-hybrid}, or introducing additional connections between the encoder and decoder to enforce the dependence between $x$ and $z$ \cite{Zhao2017InfoVAEIM, DBLP:conf/nips/GoyalSCKB17,Dieng2018AvoidingLV}; (ii) Using more flexible or multimodal priors~\cite{DBLP:journals/corr/TomczakW17, DBLP:journals/corr/abs-1808-10805}; (iii) Alternating the training by focusing on the inference network in the earlier stages~\cite{he2018lagging}, or augmenting amortized optimization of VAEs with instance-based optimization of stochastic variational inference~\cite{Kim2018SemiAmortizedVA,marino2018iterative}. 

All of the aforementioned approaches impose one or more of the following limitations: restraining the choice of decoder, modifying the training algorithm, or requiring a substantial alternation of the objective function. As exceptions to these, $\delta$-VAE~ \cite{razavi2018preventing} and $\beta$-VAE ~\cite{Matthey2017betaVAELB}  aim to avoid the posterior collapse by explicitly controlling the regularizer term in eqn.~\ref{elboobjective}. While $\delta$-VAE aims to impose a lower bound on the divergence term, $\beta$-VAE (\cref{betavae}) controls the impact of regularization via an additional hyperparameter (i.e., $\beta D_{KL}\big(q_\phi({z}|{x}) || p({z})\big)$). A special case of $\beta$-VAE is  annealing~\cite{DBLP:journals/corr/BowmanVVDJB15}, where $\beta$ increases from 0 to 1 during training. 

In this study, we propose to use an extension of $\beta$-VAE ~\cite{article123} which permits us to explicitly control the magnitude of the KL term while avoiding the posterior collapse issue even in the existence of a powerful decoder. We use this framework to examine different properties of the estimated posterior and the generative behaviour of VAEs and discuss them in the context of text generation via various qualitative and quantitative experiments.

\section{ Kullback-Leibler Divergence in VAE}\label{ELBO}
We take the encoder-decoder of VAEs as the sender-receiver in a communication network. Given an input message $x$, a sender generates a compressed encoding of $x$ denoted by $z$, while the receiver aims to fully decode $z$ back into $x$. The quality of this communication can be explained in terms of \emph{rate}~(R) which measures the compression level of $z$ as compared to the original message $x$, and \emph{distortion}~(D) which quantities the overall performance of the communication in encoding a message at sender and successfully decoding it at the receiver. Additionally, the capacity of the encoder channel can be measured in terms of the amount of mutual information between $x$ and $z$, denoted by $\text{I}({x};{z})$~\cite{cover2012elements}.

\subsection{Reconstruction vs. KL }\label{expl}
The reconstruction loss can naturally measure distortion ($D := - \big \langle \log p_\theta({x}|{z}) \big \rangle$), while the KL term quantifies the amount of compression (rate; $R := D_{KL}[q_\phi({z}|{x})|| p({z})]$) by measuring the divergence between a channel that transmits zero bit of information about $x$, denoted by $p(z)$, and the encoder channel of VAEs, $q_\phi(z|x)$. \noindent\citet{DBLP:journals/corr/abs-1711-00464} introduced the
$H-D \leq \text{I}({x};{z}) \leq R$ bounds\footnote{This is dependent on the choice of encoder. For other bounds on mutual information see~\citet{poole2018variational,ELBO_Surgery}.}, where $H$ is the empirical data entropy (a constant). 
These bounds on mutual information allow us to analyze the trade-off between the reconstruction and KL terms in eqn. (\ref{elboobjective}). For instance, since $\text{I}({x};{z})$ is non-negative (using Jensen's inequality),  the posterior collapse can be explained as the situation where $\text{I}({x};{z})=0$, where encoder transmits no information about $x$, causing $R=0, D=H$.
Increasing $\text{I}({x};{z})$ can be encouraged by increasing both bounds: increasing the upper-bound (KL term) can be seen as the mean to control the maximum capacity of the encoder channel, while reducing the distortion (reconstruction loss) will tighten the bound by pushing the lower bound to its limits ($H-D\rightarrow H$). A similar effect on the lower-bound can be encouraged by using stronger decoders which could potentially decrease the reconstruction loss. Hence, having a framework that permits the use of strong decoders while avoiding the posterior collapse is desirable. Similarly, channel capacity can be decreased.
\subsection{Explicit KL Control via $\beta$-VAE}\label{betavae}
Given the above interpretation, we now turn to a slightly different formulation of ELBO based on $\beta$-VAE~\cite{Matthey2017betaVAELB}. This allows control of the trade-off between the reconstruction and KL terms, as well as to set explicit KL value. While $\beta$-VAE offers regularizing the ELBO via an additional coefficient $\beta \in {\rm I\!R}^+$, a simple extension~\cite{article123} of its objective function incorporates an additional hyperparameter $C$ to explicitly control the magnitude of the KL term,
\begin{equation}
\footnotesize
\setlength{\abovedisplayskip}{-5pt}
\setlength{\belowdisplayskip}{4pt}
\label{betaobj}
    \big \langle\! \log p_\theta({x}|{z}) \big \rangle_{q_\phi({z}|{x})}  -  \beta |D_{KL}\big(q_\phi({z}|{x}) || p({z})\big)-C|
\end{equation}
where $C\!\!  \in\!\! {\rm I\!R}^+$ and $| . |$ denotes the absolute value. While we could apply constraint optimization to impose the explicit constraint of $\text{KL}\!\!=\!\!C$, we found that the above objective function satisfies the constraint (\cref{experiment}). Alternatively, it has been shown~\cite{pelsmaeker2019effective} the similar effect could be reached by replacing the second term in eqn.~\ref{betaobj} with $\max\big(C,D_{KL}\big(q_\phi({z}|{x}) || p({z})\big)\big)$ at the risk of breaking the ELBO when $\text{KL}\!\!<\!\!C$~\cite{NIPS2016_6581}.

\section{Experiments}\label{experiment}
We conduct various experiments to illustrate the properties that are encouraged via different KL magnitudes. In particular, we revisit the interdependence between rate and distortion, and shed light on the impact of KL on the sharpness of the approximated posteriors. Then, through a set of qualitative and quantitative experiments for text generation, we demonstrate how certain generative behaviours could be imposed on VAEs via a range of maximum channel capacities. Finally, we run some experiments to find if any form of syntactic information is encoded in the latent space. For all experiments, we use the objective function of eqn.~\ref{betaobj} with $\beta=1$. We do not use larger $\beta$s because the constraint $\text{KL}=C$ is always satisfied.~\footnote{$\beta$ can be seen as a Lagrange multiplier and any $\beta$ value that allows for constraint satisfaction ($R = C$) is fine.} 
\begin{figure}[t]
 \centering
  \includegraphics[trim={0cm 0cm -0.5cm 0cm}, width=8cm, height=10cm]{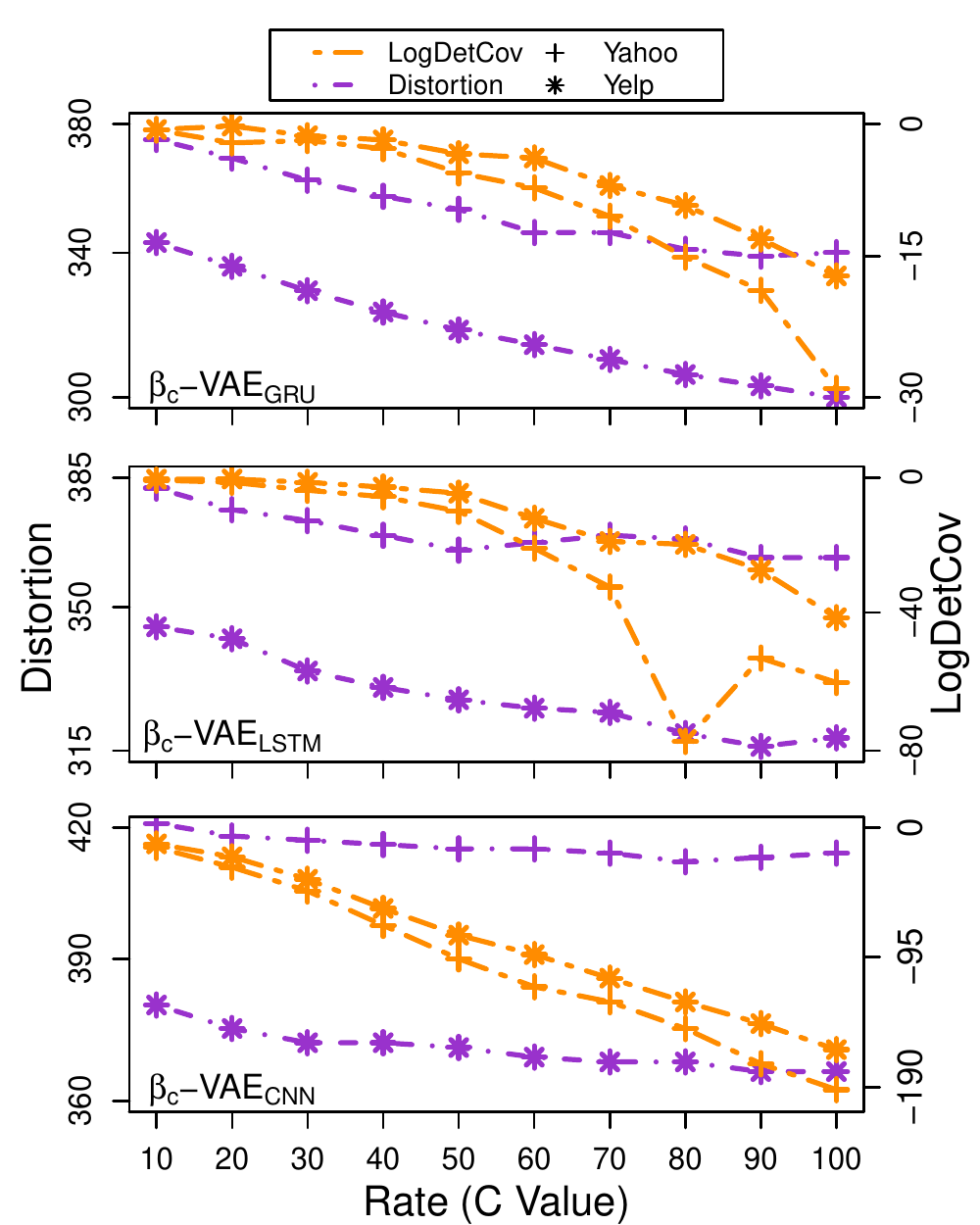}
  \caption{Rate-Distortion and \emph{LogDetCov} for $C=\{10,20,...,100\}$ on Yahoo and Yelp corpora.}
  \label{fig:sub1}
\end{figure}
\paragraph{Corpora} We use 5 different corpora covering different domains and size through this section: Yelp and Yahoo~\citet{DBLP:journals/corr/YangHSB17} both have ($100k$,$10k$,$10k$) sentences in (train, dev, test) sets and $20k$ words in vocabulary,  Children's Book Test (CBT; \citet{Weston2016TowardsAQ}) has ($192k$,$10k$,$12k$) sentences and $12k$ vocab, Wikipedia (WIKI; \citet{DBLP:journals/corr/abs-1808-09031}) has ($2m$,$270k$,$270k$) sentences and $20k$ vocab, and WebText \cite{radford2019language} has ($1m$,$23k$,$24k$) sentences and $22k$ vocab.~\footnote{ Corpora and preprocessing scripts will be released.}

\paragraph{Models} We examine three VAE architectures, covering a range of decoding strengths to examine if the objective function in eqn.~\ref{betaobj} is immune to posterior collapse regardless of the choice of encoder-decoder architectures: $\beta_C$-VAE\textsubscript{LSTM} with (LSTM encoder, LSTM decoder), $\beta_C$-VAE\textsubscript{GRU}  with (GRU encoder, GRU decoder)~ \cite{DBLP:journals/corr/ChoMGBSB14}, and $\beta_C$-VAE\textsubscript{CNN} with (LSTM encoder, CNN decoder)~\cite{DBLP:journals/corr/DauphinFAG16}. The dimension of word embeddings is 256 and the dimension of the latent variable is 64. The encoder and the decoder, for both VAE\textsubscript{LSTM} and VAE\textsubscript{GRU}, have hidden size of 512 dimensions. VAE\textsubscript{CNN} has exactly the same encoder as VAE\textsubscript{LSTM}, while the decoder follows similar architecture to GLU with a bottleneck structure (with two blocks)~ \cite{DBLP:journals/corr/DauphinFAG16} and has 512 channels externally and 128 internally for the convolutions with the filter size of 20. All models were trained for 10 epochs and optimised the objective function (eqn.~\ref{betaobj}) with Adam \cite{Kingma2015AdamAM} with following learning rates: $10^{-5}\times 85$ for VAE\textsubscript{GRU} and VAE\textsubscript{LSTM}, and $10^{-4}$ for VAE\textsubscript{CNN}. To couple the encoder with the decoder we concatenate the latent variable to word embeddings at each time step without initialisation of hidden state. 

\begin{table*}[t]
\setlength{\tabcolsep}{3pt}
\centering
\scalebox{0.70}{
\begin{tabular}{lc: ccccc : cccccccc }
\toprule
  &&&&&&&\multicolumn{2}{c}{\bf Bucket 1}  & \multicolumn{2}{c}{\bf Bucket 2}  &
 \multicolumn{2}{c}{\bf Bucket 3}  & \multicolumn{2}{c}{\bf All}   \\
\cmidrule(lr){8-9}
\cmidrule(lr){10-11}
\cmidrule(lr){12-13}
\cmidrule(lr){14-15}
&C& D  & R  & LogDetCov & $||\mu||^2_2$ & AU &  BL2/RG2& BL4/RG4 &BL2/RG2& BL4/RG4 & BL2/RG2& BL4/RG4 & BL2/RG2& BL4/RG4   \\
 \midrule
 \bf \parbox[t]{2mm}{\multirow{3}{*}{\rotatebox[origin=c]{90}{CBT}}} &3& 62  &   3 &-0.39 & 0.05& 8 & 7.49/2.63& 1.28/0.13 &8.95/3.48& 1.49/0.10 &10.37/4.81& 1.68/0.12 &9.46/3.54& 1.57/0.12 \\
 &15& 53 &   15 &-0.38 & 0.05& 29 &21.68/12.92& 8.99/3.07 &14.82/7.01& 4.25/0.81 &14.68/6.73& 3.31/0.36 &15.87/8.86& 4.60/1.43  \\
 &100 &32 & 99 &-43.82& 1.27 &64& 50.00/43.23& 38.74/30.16 &26.78/18.49& 15.99/9.23 &19.24/9.90& 7.65/2.24 &27.65/24.33& 16.09/14.24\\
  \midrule
 \bf \parbox[t]{2mm}{\multirow{3}{*}{\rotatebox[origin=c]{90}{WIKI}}}&3 &81 & 3  &-0.35& 0.00& 5&4.61/3.64& 1.47/1.03  &5.93/2.67& 1.09/0.19  &7.39/3.00& 1.17/0.12 &6.78/3.08& 1.33/0.42 \\
 &15 &70 & 15 &-0.57& 0.01& 12 &13.73/8.46& 7.12/3.86  &10.07/4.45& 3.93/1.32  &9.93/3.27& 1.95/0.29 &10.08/5.35& 3.42/1.79   \\
 &100 &17 & 100 &-4.97& 0.15& 64 &65.67/63.17& 60.02/55.92 &37.25/32.76& 30.88/26.33 &18.73/11.41& 11.22/6.20 &31.84/35.37& 24.17/29.08 \\
  \midrule
\bf \parbox[t]{2mm}{\multirow{3}{*}{\rotatebox[origin=c]{90}{WebText}}}&3 &77& 3 &-0.21 &0.01 &4 &9.51/5.27& 2.96/1.14 &9.59/4.59& 1.68/0.22 &12.59/6.37& 3.96/1.01 & 11.88/5.54& 3.35/0.70 \\
&15&67& 15 &-0.51& 0.01& 16 &21.69/12.41& 9.86/3.69  &15.48/7.44& 5.35/1.51 & 15.63/7.29& 5.59/1.59 & 15.84/7.85& 5.69/1.76   \\
&100&22& 100 &-7.85& 0.41& 64&84.85/82.48& 81.89/78.79 &61.65/58.33& 56.35/53.05  &35.07/27.33& 27.31/20.99 & 45.84/45.30& 38.71/39.66 \\
 \bottomrule
\end{tabular}
}
\caption{$\beta_C$-VAE\textsubscript{LSTM} performance with $C=\{3, 15, 100\}$ on the test sets of CBT,  WIKI, and  WebText. Each bucket groups sentences of certain length.  Bucket 1:  length $\leq$ 10; Bucket 2:  10 $<$ length $\leq$ 20; Bucket 3:  20 $<$ length $\leq$ 30, and \textbf{All} contains  all sentences. BL2/RG2 denotes BLEU-2/ROUGE-2, BL4/RG4 denotes BLEU-2/ROUGE-2 BLEU-4/ROUGE-4, AU denotes active units, D denotes distortion, and R denotes rate.}
\label{table:test_set}
\end{table*}
\subsection{Rate and Distortion}\label{rateanddistrotion}
To analyse the dependence between the values of explicit rate ($C$) and distortion,  we trained our models with different values of $C$, ranging from 10 to 100. Figure \ref{fig:sub1} reports the results for $\beta_C$-VAE\textsubscript{GRU}, $\beta_C$-VAE\textsubscript{LSTM}, and $\beta_C$-VAE\textsubscript{CNN} models on Yahoo and Yelp corpora. 
In all our experiments we found that $C\!-\!1\!\leq KL\!\leq\! C\!+\!1$, demonstrating that the objective function effectively imposed the desired constraint on KL term. Hence, setting any $C>0$ can in practice avoid the collapse issue.

The general trend is that by increasing the value of $C$ one can get a better reconstruction (lower distortion) while the amount of gain varies depending on the VAE's architecture and corpus.~\footnote{We attribute the difference in performance across our models to the non-optimal selection of training hyperparameters, and corpus specific factors such as sentence length.} Additionally, we measured rate and distortion on CBT, WIKI, and WebText corpora using $\beta_C$-VAE\textsubscript{LSTM} and observed the same trend with the increase of $C$, see Table~\ref{table:test_set}. This observation is consistent with the bound on $\text{I}({x};{z})$ we discussed  earlier (\cref{expl}) such that with an increase of KL we increase an upper bound on  $\text{I}({x};{z})$ which in turn allows to have smaller values of reconstruction loss. Additionally, as reported in Table~\ref{table:test_set}, encouraging higher rates (via larger $C$) encourages more active units (AU; \citet{Burda2015ImportanceWA}) in the latent code $z$.~\footnote{To see if the conclusions hold with different number of parameters, we doubled the number of parameters in $\beta_C$-VAE\textsubscript{GRU} and  $\beta_C$-VAE\textsubscript{LSTM} and observed the similar pattern with a slight change in performance.}

As an additional verification, we also group the test sentences into buckets based on their length and report BLEU-2/4 and ROUGE-2/4 metrics to measure the quality of reconstruction step in Table \ref{table:test_set}. As expected, we observe that increasing rate has a consistently positive impact on improving BLEU and ROUGE scores.
\subsection{Aggregated Posterior}\label{section:aggegated_posterior}
To understand how the approximated posteriors are being affected by the magnitude of the KL, we adopted an approach from \citet{Zhao2017InfoVAEIM} and looked at the divergence between the aggregated posterior, $q_\phi(z)=\sum_{x\sim q(x)} q_\phi(z|x)$, and prior $p(z$). Since during generation we generate samples from the prior, ideally we would like the aggregated posterior to be as close as possible to the prior.

We obtained unbiased samples of ${z}$ first by sampling an ${x}$ from data and then ${z} \sim q_\phi({z}|{x})$, and measured the log determinant of covariance of the samples ($\log \det(\mathrm{Cov}[q_\phi({z})])$).
As reported in Figure \ref{fig:sub1}, we observed that $\log \det(\mathrm{Cov}[q_\phi({z})])$ degrades as $C$ grows, indicating sharper approximate posteriors. We then consider the difference of $p(z)$ and $q(z)$ in their means and variances,
by computing the KL divergence from the moment-matching Gaussian fit of $q(z)$ to $p(z)$:
This returns smaller values for $\beta_{C=5}$-VAE\textsubscript{GRU} (Yelp: 0, Yahoo: 0), and larger values for $\beta_{C=100}$-VAE\textsubscript{GRU} (Yelp: 8, Yahoo: 5), which illustrates that the overlap between $q_\phi({z})$ and $p(z)$ shrinks further as $C$ grows.
 
 The above observation is better pronounced in Table~\ref{table:test_set}, where we also report the mean ($||\mu||^2_2$) of unbiased samples of $z$, highlighting the divergence from the mean of the prior distribution as rate increases. Therefore, for the case of lower $C$, the latent variables observed during training are closer to the generated sample from the prior which makes the decoder more suitable for generation purpose. We will examine this hypothesis in the following section.

\begin{table*}[t]
\setlength{\tabcolsep}{3.5pt}
\scalebox{0.71}{
\begin{tabular}{p{1.3cm} | p{6.7cm} | p{6.7cm }| p{6.7cm} }
\toprule
& \multicolumn{1}{c|}{\bf Greedy} &  \multicolumn{1}{c|}{\bf Top-15} & \multicolumn{1}{c}{ \bf NS(p=0.9)}\\
\midrule
\multirow{9}{*}{\bf C=3}& 1: oh, i m not going to be a good man. & 1: come - look on my mind, said he. & 1: and what is one of those trees creatures? \\
&2: oh, it s a good thing, said the story girl. & 2: how could i tell you, that it s a great deal? & 2: here s a nice heart among those waters!\\
&3:\cellcolor[HTML]{C0C0C0}oh, how can you do it, dear? & 3: said i. my sister, what a fool! & 3: good-bye, said reddy fox, hardly frightened was out of his life.\\
&4: \cellcolor[HTML]{C0C0C0} oh, how can you do it, dear? & 4: and how was the way, you? & 4: now, for a neighbor, who knows him. \\
&5: oh, how can you do it, miss? & 5: said the other little breezes, but i do n t . & 5: oh, prince ivan, dear me!\\
&6:\cellcolor[HTML]{C0C0C0} and what is the matter with you? & 6: and where s the news of the world? & 6: cried her mother, who is hidden or power. \\
&7: \cellcolor[HTML]{C0C0C0} and what is the matter with you? & 7: \textlangle{}unk\textrangle{} of \textlangle{}unk\textrangle{}, said i. ay, \textlangle{}unk\textrangle{}! & 7: but this was his plight, and the smith knew.\\

\midrule
\multirow{13}{*}{\bf C=15}& 1: old mother west wind and her eyes were in the same place, but she had never seen her. &  1: eric found out this little while, but there in which the old man did not see it so.& 1: aunt tommy took a sudden notion of relief and yellow-dog between him sharply until he tried to go to. \\
&2: old mother west wind and his wife had gone and went to bed to the palace. & 2: old mother west wind and his wife gave her to take a great \textlangle{}unk\textrangle{}, she said. & 2: his lord marquis of laughter expressed that soft hope and miss cornelia was not comforted.\\
&3: little joe otter and there were a \textlangle{}unk\textrangle{} of them to be seen. & 3: little joe otter got back to school all the \textlangle{}unk\textrangle{} together. & 3: meanwhile the hounds were both around and then by a thing was not yet.\\
&4: little joe otter s eyes are just as big as her. & 4: little joyce s eyes grew well at once, there. &  4: in a tone, he began to enter after dinner.\\
&5: a few minutes did not answer the \textlangle{}unk\textrangle{}. & 5: pretty a woman, but there had vanished. & 5: once a word became, just got his way.\\
&6: a little while they went on. & 6: from the third day, she went. & 6: for a few moments, began to find.\\
&7: a little while they went. & 7: three months were as usual. & 7: meantime the thrushes were \textlangle{}unk\textrangle{}.\\

\midrule
\multirow{13}{*}{\bf C=100} &1: it will it, all her \textlangle{}unk\textrangle{}, not even her with her? & 1: it will her you, at last, bad and never in her eyes. & 1: it s; they liked the red, but i kept her and growing. \\
&2: it will get him to mrs. matilda and nothing to eat her long clothes. & 2: other time, i went into a moment -- she went in home and. & 2:  it \textlangle{}unk\textrangle{} not to her, in school, and never his bitter now.\\
&3: the thing she put to his love, when it were \textlangle{}unk\textrangle{} and too. & 3: going quite well to his mother, and remember it the night in night! & 3: was it now of the beginning, and dr. hamilton was her away and.\\
&4: one day, to the green forest now and a long time ago, sighed. & 4: one and it rained for his feet, for she was their eyes like ever. & 4: of course she flew for a long distance; and they came a longing now. \\
&5: one and it became clear of him on that direction by the night ago. & 5: the thing knew the tracks of \textlangle{}unk\textrangle{} and he never got an \textlangle{}unk\textrangle{} before him. & 5: one door what made the pain called for her first ear for losing up.\\
&6: every word of his horse was and the rest as the others were ready for him. & 6: of course he heard a sound of her as much over the \textlangle{}unk\textrangle{} that night can. & 6: one and he got by looking quite like her part till the marriage know ended. \\
&7: a time and was half the \textlangle{}unk\textrangle{} as before the first \textlangle{}unk\textrangle{} things were ready as. &7: every, who had an interest in that till his legs got splendid tongue than himself. & 7: without the thought that danced in the ground which made these delicate child s teeth so.\\
\bottomrule
\end{tabular}
}
\caption{Homotopy (CBT corpus) -  The three blocks correspond to $C=\{3,15,100\}$ values used for training $\beta_{C}$-VAE\textsubscript{LSTM}. The columns correspond to the three decoding schemes: greedy, top-k (with k=15), and the nucleus sampling (NS; with p=0.9). Initial two latent variables $z$ were sampled from a the prior distribution i.e. $z\sim p(z)$ and the other five latent variables were obtained by interpolation. The sequences that highlighted in gray are the one that decoded into the same sentences condition on different latent variable. \textbf{Note}: Even though the learned latent representation should be quite different for different models (trained with different C) in order to be consistent all the generated sequences presented in the table were decoded from the same seven latent variables. } 
\label{table:homotopy}
\end{table*}
 \subsection{Text Generation}
To empirically examine how channel capacity translates into generative capacity of the model, we experimented with the $\beta_C$-VAE\textsubscript{LSTM} models from Table~\ref{table:test_set}. To generate a novel sentence, after a model was trained, a latent variable $z$ is sampled from the prior distribution and then transformed into a sequence of words by the decoder $p(x|z)$. 

During decoding for generation we try three decoding schemes: (i) Greedy: which selects the most probable word at each step, (ii) Top-k~\cite{DBLP:journals/corr/abs-1805-04833}: which at each step samples from the K most probable words, and (iii) Nucleus Sampling (NS)~ \cite{DBLP:journals/corr/abs-1904-09751}: which at each step samples from a flexible subset of most probable words chosen based on their cumulative mass (set by a threshold $p$, where $p = 1$ means sampling from the full distribution). While similar to Top-k, the benefit of NS scheme is that the vocabulary size at each time step of decoding varies, a property that encourages diversity and avoids degenerate text patterns of greedy or beam search decoding~\cite{DBLP:journals/corr/abs-1904-09751}. We experiment with NS $(p=\{0.5, 0.9\})$ and Top-k $(k=\{5, 15\})$.

\subsubsection{Qualitative Analysis} 
We follow the settings of homotopy experiment~\cite{DBLP:journals/corr/BowmanVVDJB15} where first a set of latent variables was obtained by performing a linear interpolation between $z_1 \sim p(z)$ and $z_2 \sim p(z)$. Then each $z$ in the set was converted into a sequence of words by the decoder $p(x|z)$. 
Besides the initial motivation of \citet{DBLP:journals/corr/BowmanVVDJB15} to examine how neighbouring latent codes look like, our additional incentive is to analyse how sensitive the decoder is to small variations in the latent variable when trained with different channel capacities,  $C=\{3,15,100\}$.

\begin{table*}[t!]
\setlength{\tabcolsep}{10.5pt}
\centering
\scalebox{0.8}{
\begin{tabular}{lc: ccccc: ccccc }
\toprule
 
&&\multicolumn{5}{c}{\bf Greedy}   & \multicolumn{5}{c}{\bf NS(p=0.9)} \\

\cmidrule(lr){3-7}
\cmidrule(lr){8-12}

&C& $|$V$|$ & FCE & \%unk & len. & SB
& $|$V$|$ & FCE & \%unk & len. & SB
 \\
 \midrule
 \bf \parbox[t]{2mm}{\multirow{3}{*}{CBT}} &3& 335  & 86.6(0.4) & 9.7 & 15.3 &4.2 &

9.8k   & 70.4(0.0) & 2.1 & 15.6 &0.0   \\
 
&15& 335  & 52.3(0.3) & 12.7 &  15.2 &0.3 &
  
 9.8k   & 70.7(0.2)& 2.4 & 15.4 &0.0 \\

&100& 335  & 47.3(0.1) & 21.3 &  17.5 &0.0 &
 
 9.8k  & 75.1(0.1) & 2.2 & 17.6 & 0.0  \\\hdashline

Test& & 328  & - & 30.7 &  15.3 & - &

6.1k  & - & 3.6  & 15.3 &- \\
 \midrule

\bf \parbox[t]{2mm}{\multirow{3}{*}{WIKI}} &3& 1.5k  & 134.6(0.8) & 27.3 &  19.9 &7.6&
  
 20k  & 89.8(0.1)& 5.8 & 19.4& 0.0  \\

&15& 1.5k & 69.2(0.1) & 18.9 & 19.8 &0.2&
  
20k & 89.3(0.1) & 5.6  & 19.8 & 0.0  \\
 
&100& 1.5k  & 58.9(0.1) & 34.8 &  20.7 & 0.0&
 
 20k  & 96.5(0.1) & 4.5 & 20.7 & 0.0  \\\hdashline
 
Test&& 1.5k  & - & 32.7 &  19.6 &- &

20k  & - & 5.2 & 19.6 &-  \\
\midrule

 \bf \multirow{3}{*}{WebText} &3& 2.3k  & 115.8(0.7) & 18.8 &  17.5 &2.0 &
 
21.9k  & 86.4(0.1) & 7.1 & 15.6 &0.0  \\
 
&15& 2.3k  & 74.4(0.1) & 15.5  & 15.8 &0.1 &
 
 21.9k  & 85.8(0.1)& 6.9 & 15.9 &0.0  \\

&100& 2.3k  & 62.5(0.1) & 27.3 & 18.0 &0.0 &

 21.9k  & 93.7(0.1) & 4.8 & 18.0 &0.0  \\\hdashline
 
Test&& 2.2k  & - & 30.1 & 16.1 &- &

 17.1k & - & 6.8 & 16.1 &- \\
\bottomrule
\end{tabular}
}
\caption{Forward Cross Entropy (FCE). Columns represent stats for Greedy and NS decoding schemes for $\beta_C$-VAE\textsubscript{LSTM} models trained with  $C=\{3,15,100\}$ on CBT, WIKI or WebText. Each entry in the table is a mean of negative log likelihood of an LM. The values in the brackets are the standard deviations. $|$V$|$ is the vocabulary size; Test stands for test set; \%unk is the percentage of \textlangle{}unk\textrangle{} symbols in a corpora; len. is the average length of a sentence in the generated corpus; SB is the self-BLEU:4 score calculated on the 10K sentences in the generated corpus.}
\label{table:FCE}
\end{table*}

Table \ref{table:homotopy} shows the generated sentences via different decoding schemes for each channel capacity. For space reason, we only report the generated sentences for greedy, Top-$k=15$, and NS $p=0.9$. To make the generated sequences comparable across different decoding schemes or C values, we use the same samples of $z$ for decoding.

\paragraph{Sensitivity of Decoder} To examine the sensitivity\footnote{Note: we vary z in one (randomly selected) direction (interpolating between $z_1$ and $z_2$). Alternatively, the sensitivity analysis can be done by varying $z$ along the gradient direction of $\log p(x | z)$.} of the decoder to variations of the latent variable, we consider the sentences generate with the greedy decoding scheme (the first column in Table \ref{table:homotopy}). The other two schemes are not suitable for this analysis as they include sampling procedure. This means that if we decode the same latent variable twice we will get two different sentences.  We observed that with lower channel capacity ($C=3$) the decoder tends to generate identical sentences for the interpolated latent variables (we highlight these sentences in gray), exhibiting decoder's lower sensitivity to $z$'s variations. However, with the increase of channel capacity ($C=15,100$) the decoder becomes more sensitive. This observation
is further supported by the increasing pattern of active units in Table \ref{table:test_set}: Given that AU increases with increase of $C$ one would expect that activation pattern of a latent variable becomes more complex as it comprises more information. Therefore small change in the pattern would have a greater effect on the decoder.


\paragraph{Coherence of Sequences} We observe that the model trained with large values of $C$ compromises sequences' coherence during the sampling. This is especially evident when we compare $C=3$ with $C=100$. Analysis of Top-15 and NS~(p=0.9) generated samples reveals that the lack of coherence is not due to the greedy decoding scheme per se, and can be attributed to the model in general. 
To understand this behavior further,
we need two additional results from Table \ref{table:test_set}: LogDetCov and $||\mu||^2_2$. One can notice that as $C$ increases LogDetCov decreases and $||\mu||^2_2$ increases. This indicates that the aggregated posterior becomes further apart from the prior, hence the latent codes seen during the training diverge more from the codes sampled from the prior during generation. We speculate this contributes to the coherence of the generated samples, as the decoder is not equipped to decode prior samples properly at higher $C$s.

\subsubsection{Quantitative Analysis} 
Quantitative analysis of generated text without gold reference sequences (e.g. in Machine Translation or Summarization) has been a long-standing challenge.
Recently, there have been efforts towards this direction, with proposal such as self-BLEU \cite{Zhu:2018:TBP:3209978.3210080}, forward cross entropy  \cite[FCE]{DBLP:journals/corr/abs-1804-07972} and Fr\'{e}chet InferSent Distance  \cite[FID]{DBLP:journals/corr/abs-1804-07972}.
We opted for FCE as a complementary metric to our qualitative analysis. 
%
%
To calculate FCE, first 
a collection of synthetic sentences are generated by sampling $z\sim p(z)$ and decoding  the samples into sentences. The synthetic sequences are then used to train a language model (an LSTM with the parametrisation of our decoder). The FCE score is estimated by reporting the negative log likelihood (NLL) of the trained LM on the set of human generated sentences.

%
We generated synthetic corpora using trained models from Table~\ref{table:test_set} with different C and decoding schemes and using the same exact $z$ samples for all corpora. Since the generated corpora using different C values would have different coverage of words in the test set (i.e., Out-of-Vocabulary ratios), we used a fixed vocabulary to minimize the effect of different vocabularies in our analysis. Our dictionary contains words that are common in all of the three corpora, while the rest of the words that don't exist in this dictionary are replaced with \textlangle{}unk\textrangle{} symbol. Similarly, we used this fixed dictionary to preprocess the test sets.  Also, to reduce bias to a particular set of sampled $z$'s we measure the FCE score three times, each time we sampled a new training corpus from a $\beta_C$-VAE\textsubscript{LSTM} decoder and trained an LM from scratch. In Table \ref{table:FCE} we report the average FCE (NLL) for the generated corpora. 

In the qualitative analysis we observed that the text generated by the $\beta_C$-VAE\textsubscript{LSTM} trained with large values of $C=100$ exhibits lower quality (i.e., in terms of coherence).
This observation is supported by the FCE score of NS(p=0.9) decoding scheme (\ref{table:FCE}), since the performance drops when the LM is trained on the corpus generated with $C=100$. The generated corpora with $C=3$ and $C=15$ achieve similar FCE score. However, these patterns are reversed for Greedy decoding scheme\footnote{For the other decoding schemes: Top-\{5,15\} and NS(p=0.5) the pattern is the same as for the Greedy. For space reason we only report the FCE for Greedy.}, where the general tendency of FCE scores suggests that for larger values of $C$ the $\beta_C$-VAE\textsubscript{LSTM} seems to generate text which better approximates the natural sentences in the test set. To understand this further, we report additional statistics in Table \ref{table:FCE}: percentage of \textlangle{}unk\textrangle{} symbols, self-BLEU and average sentence length in the corpus. 

The average sentence length, in the generated corpora is very similar for both decoding schemes, removing the possibility that the pathological pattern on FCE scores was caused by difference in sentence length.  However, we observe that for Greedy decoding more than $30\%$ of the test set consists of \textlangle{}unk\textrangle{}. Intuitively, seeing more evidence of this symbol during training would improve our estimate for the \textlangle{}unk\textrangle{}. As reported in the table, the $\%$unk increases on almost all corpora as $C$ grows, which is then translated into getting a better FCE score at test. 
Therefore, we believe that FCE at high $\%$unk is not a reliable quantitative metric to assess the quality of the generated syntactic corpora. Furthermore, for Greedy decoding, self-BLEU decreases when $C$ increases. This suggests that generated sentences for higher value of $C$ are more diverse. Hence, the LM trained on more diverse corpora can generalise better, which in turn affects the FCE.


In contrast, the effect the \textlangle{}unk\textrangle{} symbol has on the corpora generated with the NS(p=0.9) decoding scheme is minimal for two reasons: First, the vocabulary size  for the generated corpora, for all values of $C$ is close to the original corpus (the corpus we used to train the $\beta_C$-VAE\textsubscript{LSTM}). Second, the vocabularies of the corpora generated with three values of $C$ is very close to each other. As a result, minimum replacement of the words with the \textlangle{}unk\textrangle{} symbol is required, making the experiment to be more reflective of the quality of the generated text. Similarly, self-BLEU for the NS(p=0.9) is the same for all values of $C$. This suggests that the diversity of sentences has minimal, if any, effect on the FCE.

\subsection{Syntactic Test}
In this section, we explore if any form of syntactic information is captured by the encoder and represented in the latent codes despite the lack of any explicit syntactic signal during the training of the $\beta_C$-VAE\textsubscript{LSTM}. To train the models we used the same WIKI data set as in \citet{DBLP:journals/corr/abs-1808-09031}, but we filtered out all the sentences that are longer than 50 space-separated tokens.\footnote{We applied the filtering to decrease the training time of our models.}

We use the data set of \citet{DBLP:journals/corr/abs-1808-09031} which consists of pairs of grammatical and ungrammatical sentences to test various syntactic phenomenon. For example, a pair in subject-verb agreement category would be: (\emph{The author laughs}, \emph{The author laugh}). 
%
%
We encode both the grammatical and ungrammatical sentences into the latent codes $z^+$ and $z^-$, respectively. Then we condition the decoder on the $z^+$  and try to determine whether the decoder assigns higher probability to the grammatical sentence (denoted by $x^+$): $p(x^-|z^+) < p(x^+|z^+)$ (denoted by p\textsubscript{1} in Table \ref{table:syntactic}). We repeat the same experiment but this time try to determine whether the decoder, when conditioned on the ungrammatical code ($z^-$), still prefers to assign higher probability to the grammatical sentence: $p(x^-|z^-) < p(x^+|z^-)$ (denoted by p\textsubscript{2} in Table \ref{table:syntactic}). Table \ref{table:syntactic} shows the p\textsubscript{1} and p\textsubscript{2} for the $\beta_C$-VAE\textsubscript{LSTM} model trained with  $C=\{3,100\}$. Both the p\textsubscript{1} and p\textsubscript{2} are similar to the accuracy and correspond to how many times a grammatical sentence was assigned a higher probability.  


\begin{table*}[t!]
\setlength{\tabcolsep}{19pt}
\centering
\scalebox{0.8}{
\begin{tabular}{l cc |cc:cc}
\toprule
 &\multicolumn{2}{c}{\textbf{$C=3$} } & \multicolumn{4}{c}{\textbf{$C=100$}}\\ 
   \cmidrule(lr){2-3}
   \cmidrule(lr){4-7}
 {\bf Syntactic Categories} & p\textsubscript{1} &  p\textsubscript{2} & p\textsubscript{1} &  p\textsubscript{2} & \=p\textsubscript{1} & \=p\textsubscript{2}\\
 \midrule
 {\bf SUBJECT-VERB AGREEMENT} &  &  &  &  &  &  \\
    Simple & 0.81 & 0.81 & 1.0               &0.23&0.68&0.47\\
    In a sentential complement & 0.79&0.79    &0.98&0.14&0.69&0.48\\
    Short VP coordination & 0.74& 0.73        &0.96&0.08&0.78&0.43\\
    Long VP coordination & 0.61& 0.61         &0.97&0.06&0.55&0.47\\
    Across a prepositional phrase & 0.78&0.78 &0.97&0.07&0.62&0.49\\
    Across a subject relative clause & 0.77&0.77&0.93&0.08 &0.68&0.41\\
    Across an object relative clause & 0.69&0.69& 0.92&0.11 &0.61 &0.45\\
    Across an object relative (no that) & 0.58&0.58 & 0.94&0.09& 0.61&0.44\\
    In an object relative clause & 0.74&0.74 & 0.99&0.01& 0.60&0.45\\
    In an object relative (no that) & 0.74&0.74& 0.99&0.02 &0.61&0.46\\
 \midrule
 {\bf  REFLEXIVE ANAPHORA} &  &  &  &  &  &  \\
 Simple & 0.79&0.78  & 0.99&0.07 & 0.70&0.39\\
 In a sentential complement & 0.74&0.73& 1.00&0.00 & 0.70&0.38\\
 Across a relative clause & 0.63&0.62 & 0.99&0.03 & 0.69&0.35\\
\midrule
{\bf NEGATIVE POLARITY ITEMS}  &  &  &  &  &  &  \\
Simple & 0.42&0.33 &1.00&0.00& 0.76&0.20\\
Across a relative clause & 0.37&0.36 & 1.00&0.00 &0.98&0.02\\

\bottomrule
\end{tabular}
}
\caption{ p\textsubscript{1}: $p(x^-|z^+) < p(x^+|z^+)$ and  p\textsubscript{2}: $p(x^-|z^-) < p(x^+|z^-)$; \=p\textsubscript{1}: $p(x^-|\bar{z}^+) < p(x^+|\bar{z}^+)$ and  \=p\textsubscript{2}: $p(x^-|\bar{z}^-) < p(x^+|\bar{z}^-)$; $\beta_{C=3}$-VAE\textsubscript{LSTM} (D:103, R:3); $\beta_{C=100}$-VAE\textsubscript{LSTM} (D:39, R:101). }
\label{table:syntactic}
\end{table*}

As reported for C=3, p\textsubscript{1} and p\textsubscript{2} match in almost all cases. This is to some degree expected since lower channel capacity encourages a more dominating decoder which in our case was trained on grammatical sentences from the WIKI. On the other hand, this illustrates that despite avoiding the KL-collapse issue, the dependence of the decoder on the latent code is so negligible that the decoder hardly distinguishes the grammatical and ungrammatical inputs. 
This changes for $C=100$, as in almost all the cases the decoder becomes strongly dependent on the latent code and can differentiate between what it has seen as input and the closely similar sentence it hasn't received as the input: The decoder assigns larger probability to the ungrammatical sentence when conditioned on the $z^-$ and, similarly, larger probability to the grammatical sentence when conditioned on the $z^+$. 
 

However, the above observations neither confirm nor reject existence of grammar signal in the latent codes. We run a second set of experiments where we aim to discard sentence specific information from the latent codes by averaging the codes\footnote{Each syntactic category is further divided into sub-categories, for instance \emph{simple subject-verb agreement} We average $z$'s within each sub-categories.}
inside each syntactic category. The averaged codes are denoted by $\bar{z}^+$ and $\bar{z}^-$, and the corresponding accuracies are reported by \=p\textsubscript{1} and \=p\textsubscript{2} in Table \ref{table:syntactic}. Our hypothesis is that the only invariant factor during averaging the codes inside a  category is the grammatical property of its corresponding sentences.

%
%
As expected, due to the weak dependence of decoder on latent code, the performance of the model under $C=3$ is almost identical (not included for space limits) when comparing p\textsubscript{1} vs. \=p\textsubscript{1}, and p\textsubscript{2} vs. \=p\textsubscript{2}. However, for $C=100$ the performance of the model deteriorates. While we leave further exploration of this behavior to our future work, we speculate this could be an indication of two things: the increase of complexity in the latent code which encourages a higher variance around the mean, or the absence of syntactic signal in the latent codes. 

\section{Discussion and Conclusion}
In this paper we analysed the interdependence of the KL term in Evidence Lower Bound (ELBO) and the properties of the approximated posterior for text generation. To perform the analysis we used an information theoretic framework based on a variant of $\beta$-VAE objective, which permits explicit control of the KL term, and treats KL as a mechanism to control the amount of information transmitted between the encoder and decoder.

The immediate impact of the explicit constraint is avoiding the collapse issue ($D_{KL}=0$) by setting a non-zero positive constraint ($C\geq 0$) on the KL term ($|D_{KL}\big(q_\phi({z}|{x}) || p({z})\big)-C|$). We experimented with a range of constraints ($C$) on the KL term and various powerful and weak decoder architectures (LSTM, GRU, and CNN), and empirically confirmed that in all cases the constraint was satisfied. 

We showed that the higher value of KL encourages not only divergence from the prior distribution, but also a sharper and more concentrated approximated posteriors. It encourages the decoder to be more sensitive to the variations on the latent code, and makes the model with higher KL less suitable for generation as the latent variables observed during training are farther away from the prior samples used during generation. To analyse its impact on generation we conducted a set of qualitative and quantitative experiments. 

In the qualitative analysis we showed that small and large values of KL term impose different properties on the generated text: the decoder trained under smaller KL term tends to generate repetitive but mainly plausible sentences, while for larger KL the generated sentences were diverse but incoherent. This behaviour was observed across three different decoding schemes and complemented by a quantitative analysis where we measured the performance of an LSTM LM trained on different VAE-generated synthetic corpora via different KL magnitudes, and tested on human generated sentences. 

%
%
%

    Finally, in an attempt to understand the ability of the latent code in VAEs to represent some form of syntactic information, we tested the ability of the model to distinguish between grammatical and ungrammatical sentences. We verified that at lower (and still non-zero) KL the decoder tends to pay less attention to the latent code, but our findings regarding the presence of a syntactic signal in the latent code were inconclusive. We leave it as a possible avenue to explore in our future work.
%
%
%
Also, we plan to develop practical algorithms for the automatic selection of the $C$'s value, and verify our  findings under multi-modal priors and complex posteriors.

\section*{Acknowledgments}
The authors would like to thank the anonymous reviewers for their helpful suggestions. This research was supported by an EPSRC Experienced Researcher Fellowship (N. Collier: EP/M005089/1), an MRC grant (M.T. Pilehvar: MR/M025160/1) and  E. Shareghi is supported by the ERC Consolidator Grant LEXICAL (648909). We gratefully acknowledge the donation of a GPU from the NVIDIA.

\bibliography{emnlp-ijcnlp-2019}
\bibliographystyle{acl_natbib}

\end{document}